\documentclass{article} 
\usepackage{iclr2021_conference,times}
\pdfoutput=1

\usepackage{amsmath,amsfonts,bm}

\def\eqref#1{equation~\ref{#1}}

\def\1{\bm{1}}

\DeclareMathAlphabet{\mathsfit}{\encodingdefault}{\sfdefault}{m}{sl}
\SetMathAlphabet{\mathsfit}{bold}{\encodingdefault}{\sfdefault}{bx}{n}

\def\sR{{\mathbb{R}}}

\newcommand{\E}{\mathbb{E}}

\usepackage{hyperref}
\usepackage{url}
\usepackage{graphicx}
\usepackage{multirow}
\graphicspath{ {./img/} }

\usepackage{xcolor}
\definecolor{tok_bg}{rgb}{0.8,0.8,0.8}

\title{Localizing Model Behavior With Path Patching}

\author{\\ \textbf{Nicholas Goldowsky-Dill}$^1$\textbf{,} \textbf{Chris MacLeod}$^1$\textbf{,} \textbf{Lucas Sato}$^1$\textbf{,} \textbf{\&} \textbf{Aryaman Arora}$^1$ \\
$^1$Redwood Research \\
\texttt{nix@rdwrs.com, chris@rdwrs.com,} \\ 
\texttt{satojk@stanford.edu, aa2190@georgetown.edu}
}

\iclrfinalcopy 
\begin{document}

\maketitle

\begin{abstract}
Localizing behaviors of neural networks to a subset of the network's components or a subset of interactions between components is a natural first step towards analyzing network mechanisms and possible failure modes. Existing work is often qualitative and ad-hoc, and there is no consensus on the appropriate way to evaluate localization claims. We introduce path patching, a technique for expressing and quantitatively testing a natural class of hypotheses expressing that behaviors are localized to a set of paths. We refine an explanation of induction heads, characterize a behavior of GPT-2, and open source a framework for efficiently running similar experiments.
\end{abstract}

\section{Introduction}

Deep neural networks can perform many complex tasks, but our ability to reason about their behavior is limited \citep{rauker2023transparent}. Recent work has reverse engineered toy networks \citep{chughtai2023toy} such that the function of every neuron can be understood. For state of the art networks, this remains out of reach and current works focus on approximately explaining small parts of the network. For example, some behaviors can be approximately understood with reference to only an abstracted ``circuit" containing a small number of interacting components \citep{wang2022interpretability, Olsson2022IncontextLA, Olah2020ZoomIA}.

Currently, we lack guarantees that deep networks will behave robustly out of distribution. To mitigate these risks, one avenue of research is generating simplified causal abstractions of the network \citep{geiger2021causal, geiger2023causal}. Such abstractions aim to be more compact and thus easier to reason about than the network, while approximating the network's behavior sufficiently well over some range of inputs. These goals inherently trade off, but we hope that even substantially incomplete abstractions can be useful in a variety of downstream tasks such as finding adversarial examples.

\emph{Path patching} was first introduced in \citet{wang2022interpretability}, where they considered a \emph{sender} attention head that interacted with the key, query, or value inputs of one or more \emph{receiver} attention heads. By performing causal interventions, they could measure composition between the heads, or the influence of a head on the logits.

In this work, we generalize path patching to test hypotheses containing any number of paths from input to output in an arbitrary computational graph. The use of paths is strictly more expressive than considering a set of graph nodes such as neurons or attention heads. We define a precisely specified format for localization claims, enabling quantitative comparison between claims and a clear understanding of their scope.

Formally, a hypothesis is a claim that a subset of paths in a network (which we call ``important paths") mediate \citep{pearl2022direct} the relationship between input and output on a given distribution; all other paths in the network are ``unimportant paths". By expressing the network as a computational graph in a particular form and removing all contributions of unimportant paths, we arrive at an approximate abstraction of the network. Path patching can then show you how similar the behavior of the abstraction is to the original network. When path patching rejects a hypothesis, \emph{path patching attribution} shows the source of the discrepancies, allowing the researcher to iteratively refine the claim.

Our contributions are:
\begin{itemize}
\item We formalize the path patching methodology for precisely describing and testing localization claims.
\item We use path patching to test and iteratively refine hypotheses about induction heads in an attention-only transformer.
\item We formalize, test, and refine a hypothesis about a behavior of GPT-2.
\item We provide an \href{https://github.com/redwoodresearch/rust_circuit_public}{open source framework} for path patching experiments.
\end{itemize}

\section{Methodology}

\subsection{Localization}

Localization is the problem of finding which parts of the network matter for a chosen behavior; it does not consider what computations they perform. The granularity of localization can vary greatly: an individual neuron \citep{finlayson2021causal}, an individual attention head \citep{vig2020investigating}, subspaces \citep{geiger2023finding}, a composition of attention heads \citep{Olsson2022IncontextLA}, a transformer block \citep{belrose2023eliciting} or ranges of MLP layers \citep{meng2022locating}. Our framework handles all of these in a uniform way using a set of paths in a computational graph, which is strictly more expressive than treating a set of nodes such as neurons or attention heads as the atom of description.

\subsection{Choosing the Dataset}

We define a ``behavior" or object of study in terms of input-output pairs on a specific dataset. This makes the choice of dataset an essential part of the behavior. An accurate approximation on one dataset may transfer to a different dataset, but in general this is not the case. As illustrated in \citet{bolukbasi2021interpretability}, parts of the network with one behavior on specific distribution can have very different behavior on other distributions. By clearly stating the domain under consideration, we avoid treating evidence in favor of a narrow explanation as sufficient to support a broader explanation.

\subsection{Path patching with nodes as mediators}

We'll consider the simpler case where nodes are mediators first, and then build up to working with paths. Let $G$ be a function with input $x \in X$ and output $y \in Y$. Our methodology applies to arbitrary functions, but for our experiments we will focus on the forward pass function of autoregressive transformers, where $x$ would be a sequence of tokens and $y$ the next-token probabilities.

The computation of $G$ can be represented by $\mathcal{G}$, a directed acyclic graph (DAG) where nodes are functions and edges are values. We can represent the same computation at different levels of granularity by expanding or combining nodes.

\begin{figure}[ht]
\begin{center}
\includegraphics[scale=0.5]{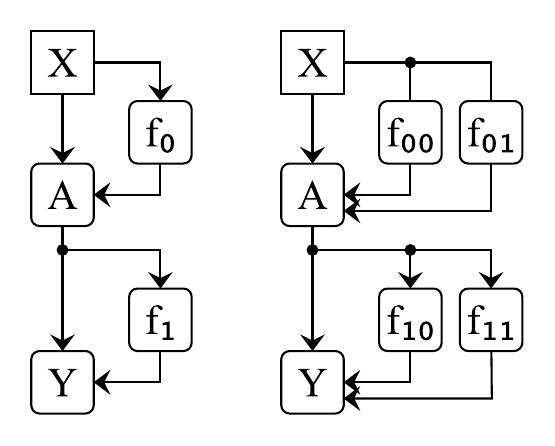}
\end{center}
\caption{Left: a two layer residual network, where $f_0$ and $f_1$ are layer functions and A and Y are skip connections that add their inputs. Right: dividing each function into two functions that sum to the original. See Appendix \ref{model_rewrites} for discussion of other common rewrites.}
\label{simple_residual}
\end{figure}

As a running example, consider a neural network with two layers and skip connections. Here $G(x) = f_1(A(x)) + A(x)$ where $A(x) = f_0(x) + x$. Two equivalent computational graphs $\mathcal{G}$ corresponding to the function $G$ are shown in Figure \ref{simple_residual}.

\begin{figure}[ht]
\begin{center}
\includegraphics[scale=0.5]{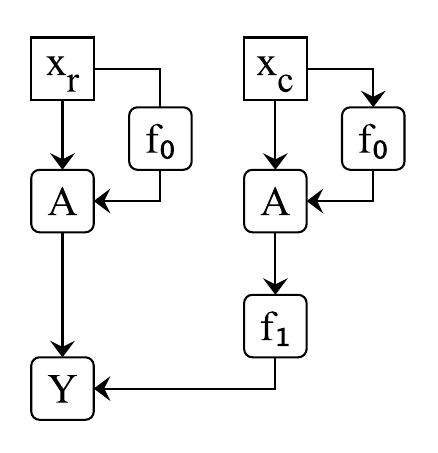}
\end{center}
\caption{To compute $G_H(x_r, x_c)$ where $f_1$ is unimportant, we replace $f_1$ in $\mathcal{G}$ with $f_1(x_c) = f_0(x_c) + x_c$. }
\label{counterfactual}
\end{figure}

Suppose we suspect that $f_1$ is an unimportant node. We define a function $G_H\colon X^2 \rightarrow Y$ where $G_H(x_r, x_c)$ means we evaluate $G$ on a \emph{reference input} $x_r$ except that we replace each unimportant node in $\mathcal{G}$ with the value that node has when evaluating $G$ on a \emph{counterfactual input} $x_c$, as shown on Figure \ref{counterfactual}. Our hypothesis predicts the output should not change. 

In the language of the causal mediation literature this measures the \emph{natural indirect effect} through $f_1$ when moving from $x_r$ to $x_c$.

Formally, a hypothesis H is a tuple $(\mathcal{G}, \delta, S, D)$ where $\delta\colon Y^2 \rightarrow \sR$ is some measure of dissimilarity, such as absolute difference between two scalars or KL divergence between two probability distributions, $S$ is a set of ``important nodes", $\mathcal{G} \backslash S$ are the unimportant nodes, and $D$ is a joint distribution over $(x_r, x_c)$ pairs. Normally the $x_r$ come from some reference distribution $D_r$ such as a training or validation set. 

There are many options for $x_c$ corresponding to different experiments: $x_c$ can come from $D_r$ (resampling), be computed via a transformation of $x_r$ (input manipulation or input corruption), be computed as the mean of $D_r$ (mean ablation of the input) or be the zero tensor (zero ablation of the input).

The strictest version of what it means for H to hold is that $\delta(G(x_r), G_H(x_r, x_c))$ is exactly zero everywhere on $D$. In modern neural networks, we find that this is too stringent and requires us to severely restrict the domain of $D$ and/or include nearly all nodes in $S$. Instead, as a notion of approximate abstraction we define the ``average unexplained effect" of H:
\begin{equation}
\label{aue_def}
\operatorname{AUE}(H) \colon= \E_{(x_r,x_c) \sim D}[\delta(G(x_r), G_H(x_r, x_c))]
\end{equation}

The hypothesis claims the AUE is zero. A nonzero AUE tells us we're missing some aspects of the behavior, which may be acceptable depending on the use case. Given two hypotheses with the same dataset and same number of paths, we generally prefer the one with a lower AUE. 

Note that this allows the unexplained effect to be very large on some inputs as long as it's small in expectation. A large unexplained effect on some inputs could indicate that a distinct mechanism is used for these inputs, in which case investigating those inputs separately would be fruitful. For applications where we care about rare failures it could be more appropriate to consider the maximum unexplained effect instead to measure worst case behavior, as in \citet{beckers2020approximate}

\subsection{Path patching with paths as mediators}

Residual networks naturally decompose into a sum of paths, where a small set of relatively shallow paths often contribute most of the effect \citep{veit2016residual}.

\begin{figure}[ht]
\begin{center}
\includegraphics[scale=0.5]{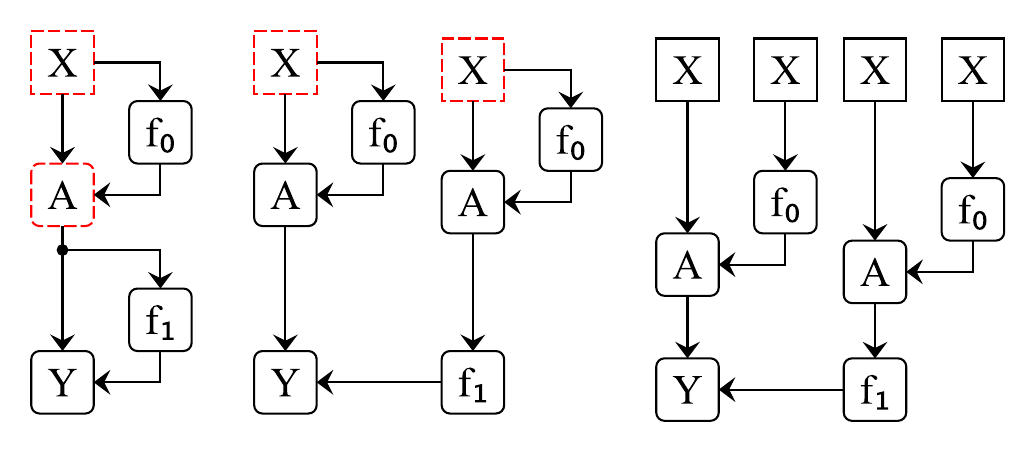}
\end{center}
\caption{$\operatorname{Treeify}(\mathcal{G})$: Left: red dashed lines are nodes with multiple outputs that will be copied. Center: Copying the subtree at A so that Y and $f_1$ take separate copies of A as input. Right: Copying X so that each A and $f_0$ has an independent copy. Note the order of copying doesn't matter; we could have duplicated X first and ended up with the same result.}
\label{gradual_treeification}
\end{figure}

In our running example, suppose our exploratory analysis suggests to us that $f_0$ and $f_1$ were both important, and that computation of $f_1$ doesn't use $f_0$'s output.  For example, this can happen in a transformer if $f_1$ is an attention layer reading from a different subspace than the subspace written by $f_0$. We express this as ``$x \rightarrow f_0 \rightarrow A \rightarrow f_1$ is unimportant". In order to feed $x_c$ to this path without affecting other paths, we introduce the function $\operatorname{Treeify}(\mathcal{G})$, which identifies each subtree that has multiple consumers of the subtree's output, and then copies the subtree so each consumer has its own copy (Figure \ref{gradual_treeification}).

The resulting graph has a one-to-one correspondence between paths in the network and copies of the input node. Now we can set the rightmost X to $x_c$ without affecting any other paths.

Algebraically, $\operatorname{Treeify}(\mathcal{G})$ implements a function $G_T\colon X^N \rightarrow Y$, where $G_T(x, x, x, \ldots , x) = G(x)$ and $N$ is the number of paths. We specify $G_T$ by fully expanding the equation for the network, then relabelling each occurrence of $x$ with an unique subscript (in arbitrary order):
\[ G_T(x_0, x_1, x_2, x_3) = f_1(A(x)) + A(x) = f_1(f_0(x_3) + x_2) + f_0(x_1) + x_0 \]

Now we allow H to be a tuple $(\mathcal{G}, \delta, P, D)$ where instead of a set of nodes S, we have a set of paths P. We define a path version of $G_H$: 
\[
G_H(x_r, x_c) \colon= G_T(x_1, x_2, \ldots, x_N) \\
\text{ with } x_i = \begin{cases}
x_r & p_i \in P \\
x_c & p_i \not \in P
\end{cases}
\]

where $p_1, p_2, \ldots p_N$ are paths through $\mathcal{G}$ in the same order as the arguments of $G_T$.

\begin{figure}[ht]
\begin{center}
\includegraphics[scale=0.5]{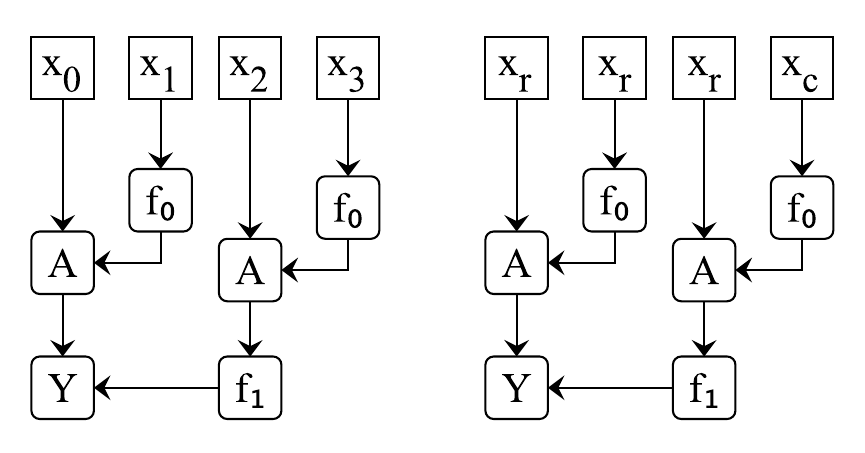}
\end{center}
\caption{Left: an arbitrary numbering of $\operatorname{Treeify}(\mathcal{G})$'s input nodes. Right: testing the hypothesis ``only $x \rightarrow f_0 \rightarrow A \rightarrow f_1$ is unimportant" reduces to testing ``only $x_3$ is unimportant".}
\label{path_counterfactual}
\end{figure}

Above, we had only one unimportant path; when there are multiple unimportant paths we simply reuse the same $x_c$ for all of them. We argue in Appendix \ref{reuse_xc} that this is a sensible default compared to other strategies such as sampling a different $x_c$ for each unimportant path.

\paragraph*{Pearl's path specific effect.}
$G_H(x_r, x_c)$ is a deterministic version of the \emph{path-specific effect} for a probabilistic graphical model (PGM). In \citet{pearl2022direct}, they modify the PGM so the input to all unselected paths is held at a reference value $x^*$. Then set the input X to the intervention value $x$ and compute the probability distribution over the output Y. If the reference $x^*$ corresponds to our $x_r$, the intervention $x$ to our $x_c$, and ``unselected paths" with our important paths, then the path-specific effect is exactly equal to $G_H(x_r, x_c)$.

\subsection{Metrics}

\paragraph*{Proportion explained.}

The \emph{average total effect} (ATE) is defined in \citet{pearl2022direct} as the effect of replacing the reference input with the counterfactual input:
\[ \operatorname{ATE(H)} \colon= \E_{(x_r, x_c) \sim D}\bigl[\delta(G(x_r), G(x_c))\bigr] \]

This is identical to the AUE of a hypothesis that no paths are important. For convenience of reporting, we can divide AUE values by the ATE to compute proportion explained as:
\[	(1 - \operatorname{AUE(H)} / \operatorname{ATE(H)}) * 100\% \]

Note that since it's possible that $\operatorname{AUE} > \operatorname{ATE}$, proportion explained can be less than 0\%. The hypothesis that all paths are important has a proportion recovered of 100\% by definition.

\paragraph*{Including the loss in the computational graph.}

Assuming we have access to ground truth labels $y\colon Y$ and a loss function $L\colon Y \times Y \rightarrow \sR$, we may wish to include the labels and loss as part of $\mathcal{G}$. When we compute an intervention, the label in the graph $y_r$ always corresponds to $x_r$ (the label is always important), $\delta$ is the absolute value function, and the unexplained effect is just the absolute difference in loss after intervention. 

We are doing this because we hope that the set of paths that are important towards getting a good loss is more sparse than the set necessary to approximate the full output distribution. In the case of cross entropy loss, only the probability assigned to the true label contributes to the loss. Any paths that only redistribute probability among the other classes are considered unimportant. For example, these paths could implement heuristics that are beneficial on a wider distribution but neutral on D.

When loss is included, it's possible to get a misleadingly high proportion recovered. For example, if the network is usually correct with high confidence, then the $\operatorname{ATE(H)} = L(G(x_c), y_r)$ will be very high and even a poor hypothesis will recover near 100\%. In this situation, a better denominator could be the loss when predicting a uniform distribution, or loss when predicting the class frequencies.

\paragraph*{Difference in expected loss.}

Suppose you were studying prompts of the form ``Which animal is bigger, [animal1] or [animal2]? The answer is:". If we consider the predicted logit difference between correct and incorrect answers, we might hope that mediators describe ``parts of the model that store animal facts". However, we would also observe mediators that are not the subject of study such as ``parts storing unigram frequencies".

If our dataset also contains symmetrical prompts with ``bigger" swapped for ``smaller", then the contribution of the unigram frequencies will cancel out on average. Swapping the order of the animals would likewise cancel out any heuristic that promotes recent tokens.

This implies we could try moving the dissimilarity metric in \eqref{aue_def} outside the expectation so that parts implementing these heuristics will test as unimportant. When we do this, our hypothesis claims the following metric will be zero:
\[ | \E_{(x_r, x_c) \sim D} [L(G_H(x_r, x_c), y_r)] - E_{x \sim D} [L(G(x), y)] | \]

We can compare to a baseline corresponding to the claim that no paths are important:
\[ | \E_{(x_r, x_c) \sim D} [L(G(x_c), y_r)] - E_{x \sim D} [L(G(x), y)] | \]

If the abstraction gets lower loss on some inputs and higher loss on other inputs, cancellation will occur and AUE will be misleadingly low \citep{scheurer2023}.

\paragraph*{Path patching attribution.}
\label{sec:path_patching_attribution}

It's often useful to take one specific $x_r$ and visualize which paths are responsible for that specific output. We call this \emph{path patching attribution} and we simply fix the joint distribution $D$ to only include that $x_r$, while sampling various $x_c$. In particular for every (prompt, completion) pair $(x_r, y_r)$ we compute:
\[ \E_{x_c \sim D} [L(G_H(x_r, x_c), y_r) - L(G(x_r), y_r)] \]

Note we drop the absolute value to preserve directional information; this allows us to see whether the patched model had a net higher or net lower loss than the original model.

\section{Results on Induction}

Given a prompt with just the token `` N", GPT-2 uses learned bigram statistics to predict common continuations like ``athan" or ``ancy". However, if given a prompt like ``Nathan and Mary went to the store. N", now GPT-2 is much more confident that ``athan" is the next token. We can represent the prompt as ``[A][B]...[A]". 

\citet{elhage2021mathematical} present a mechanistic explanation for this behavior in terms of two interacting attention heads. They note that more complex mechanisms are possible, but the minimal version is as follows. Define $i$ as the position of the second [A] and define $j$ as the position of [B]. First, a previous token head (PTH) attends from $j$ to $j-1$ and adds a vector to the residual stream representing ``[A] is previous". Second, an induction head (IH) in a later layer has a query of [A], a key of ``[A] is previous", and thus attends strongly from $i$ to $j$. Then the value operation adds a vector to the residual stream which unembeds to [B].

In this section, we apply our methodology to measure how much of the induction behavior is explained by the minimal explanation versus more complex interactions. 

\subsection{Our model}

We investigated a 2-layer attention-only autoregressive transformer trained on the OpenWebText dataset. Our model has 8 attention heads per layer preceded by layer norm, and uses shortformer positional encodings \citep{press2020shortformer}. In shortformer, learned positional embeddings are provided to the keys and queries of each attention head, but not to the values. This means positional information can be used to compute attention scores, but never enters the residual stream. The unembedding matrix is separate (in contrast to GPT-2, which has a tied unembedding).

Let the notation L.H represent an attention head, where L is the 0-indexed attention layer number, and H is the 0-indexed head number within that layer. By direct inspection of the attention patterns, we found evidence supporting head 0.0 as a PTH and heads 1.5 and 1.6 as induction heads -- see Appendix \ref{identifying_heads} for visualizations.

In our experiments, we define the dataset $D_r$ as the first 300 tokens of 100K examples from OpenWebText held-out from training. Our computational graph includes the cross-entropy loss at each individual token, which means we are explaining the probability the model places on the actual next token, without caring about the probabilities on all other tokens.

\paragraph*{Baseline}

The baseline hypothesis is that paths through the two induction heads are unimportant, and all other paths are important. That is, $x_c$ is randomly sampled from $D_r$ and used for all paths that pass through 1.5 or 1.6. Path patching shows a mean per-token absolute loss difference of 0.702 relative to the original model.

\subsection{Initial hypotheses}

As a running example we'll use a prompt from Carly Rae Jepsen's ``Call Me Maybe": 

\begin{quote}
Before you came into my life, I missed you so bad \\
I missed you so
\end{quote}

\newcommand{\tok}[1]{`\colorbox{tok_bg}{#1}'}

At the first occurrence of `` so", the model predicts `` much"\footnote{This contrasts with the much larger GPT-NeoX-20B, which has memorized the lyric and with more context can correctly predict `` bad" on the first occurrence.}, whereas at the second occurrence of `` so" it correctly predicts that `` bad" is repeated.

Our initial hypothesis claims that only the following paths are important to induction:

\begin{itemize}
\item The direct path to value hypothesis (\emph{Direct-V}) claims the value-input to the induction heads only cares about the token at $j$ via the skip connection, and we can thus patch all paths through the layer 0 heads.
\item The direct path to query hypothesis (\emph{Direct-Q}) claims the same about the query input at $i$.
\item The previous token head to key hypothesis (\emph{PTH-K}) claims the key-input to the induction heads only depends on the previous token head. At this stage, we won't claim that the PTH is only a PTH, so we'll let it use the current or any prior token.
\end{itemize}

The \emph{All-Initial} hypothesis says the union of the paths in the three hypotheses are important (a total of $1+1+13=15$ in this example).

\begin{figure}[ht]
\begin{center}
\includegraphics[scale=0.5]{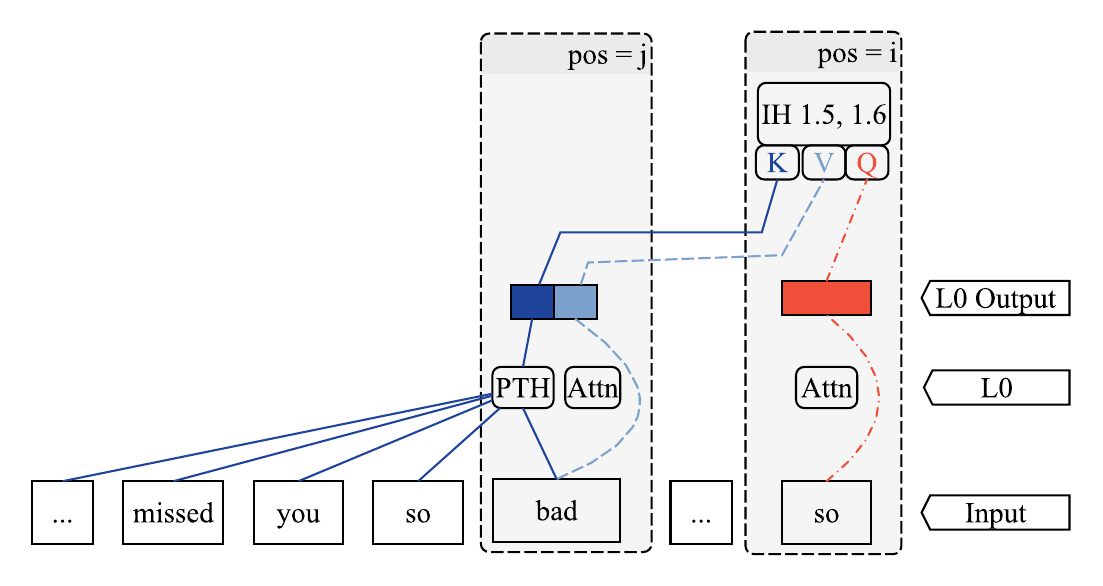}
\end{center}
\caption{PTH-K (dark blue), Direct-V (light blue, dashed), and Direct-Q (red, dot-dashed) hypotheses. In the first attention layer, only the PTH is important.}
\label{initial}
\end{figure}

Note that the positional embeddings into the Q and K inputs are not shown; at a given position these don't change when we swap tokens in an intervention, so they can be considered not important. Layer normalization (LN) before an attention head is included when the head is included. For example, in Direct-Q (above) the L0 LN is excluded and the L1 LN is included. 

The results of performing the corresponding path patching experiments are summarized below.

\begin{table}[ht]
\caption{Results of initial hypotheses}
\label{results_initial}
\begin{center}
\begin{tabular}{l|l|l|l| l}
Hypothesis & Direct-V & Direct-Q & PTH-K & All-Initial \\
\hline
Proportion explained & 64.1\% & 48.7\% & 53.3\% & 28.2\%
\end{tabular}
\end{center}
\end{table}

Recall that on this scale, 0\% means the hypothesis is as inaccurate as a hypothesis that the induction heads don't matter at all, while 100\% means that for every individual example on the dataset, the hypothesis produces equal loss to the original model.

\subsection{First refinement: Positional Hypotheses}

To the extent that a layer 0 head attends from $i$ to $i$, its output is just a linear transformation of the layer norm of the token at $i$. By visual inspection, it does appear that several attention heads have substantial attention from $i$ to $i$, implying that it's feasible for the induction heads to have adapted to reading this information in addition to reading the token embedding directly via the skip connection. 

This suggests adding 8 additional paths to the Direct-Q hypothesis of the form token $i$ $\rightarrow$ layer 0 head $\rightarrow$ induction head to form the Positional-Query Hypothesis (\emph{Positional-Q}). 

We can test this by creating a spliced input which has the last token from the reference input and all other tokens from the counterfactual input. We then compute the value of the query on this spliced input. 

We similarly define positional versions of the other two hypotheses:

\begin{itemize}
\item Positional-Value Hypothesis (\emph{Positional-V}): we add the 8 paths: token $j$ $\rightarrow$ layer 0 head $\rightarrow$ induction head.
\item Positional-Key Hypothesis (\emph{Positional-K}): inspection of the PTH's attention patterns suggests we can try eliminating tokens earlier than $j-1$ to make our hypothesis more sparse. Then we can add the paths: token $j-1$ $\rightarrow$ layer 0 head $\rightarrow$ induction head.
\end{itemize}

\begin{figure}[ht]
\begin{center}
\includegraphics[scale=0.5]{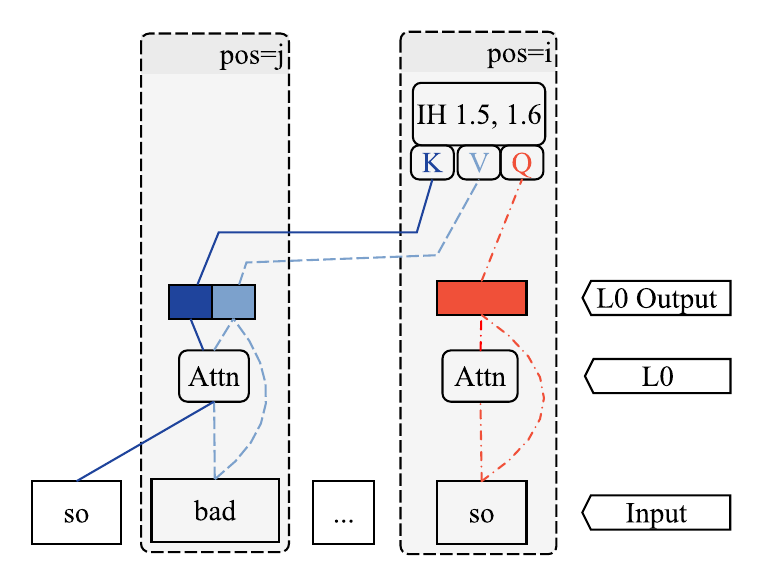}
\end{center}
\caption{The Positional-K (dark blue), Positional-V (light blue, dashed), and Positional-Q (red, dot-dashed) hypotheses.}
\label{positional}
\end{figure}

The results (with the previous numbers for comparison) are in Table \ref{results_positional}.

\begin{table}[ht]
\caption{Results of positional hypotheses}
\label{results_positional}
\begin{center}
\begin{tabular}{l l|l l|l l|l l}
\multicolumn{2}{l|}{Value} &
\multicolumn{2}{l|}{Query} & 
\multicolumn{2}{l|}{Key} &
\multicolumn{2}{l}{All} \\
\hline
Direct-V & 64.1\% & Direct-Q & 48.7\% & PTH-K & 53.5\% & All-Initial & 28.2\% \\
\hline
Positional-V & 86.2\% & Positional-Q & 72.6\% & Positional-K & 55.8\% & All-Positional & 48.6\%
\end{tabular}
\end{center}
\end{table}

\subsection{Second refinement: Long induction}

Positional-K was not significantly improved, which suggests that information earlier than $j-1$ is in fact useful. By visual inspection, it does appear that 0.0 attends almost 100\% to the previous token, but other heads (particularly 0.6) attend to multiple recent tokens. This suggests that a longer induction pattern like ``[A1][A2][B]...[A1][A2]" $\rightarrow$ [B] might be implemented in the model by a ``recent tokens head" that integrates information from both A1 and A2.

To test this, we add back paths to the key from previous tokens starting at $j-K$, and paths to the query starting at $i-K+1$. This gives the long positional query and key hypothesis (\emph{Long-QK}). We verify that $K=3$ is the shortest window that performs well.

\begin{figure}[ht]
\begin{center}
\includegraphics[scale=0.5]{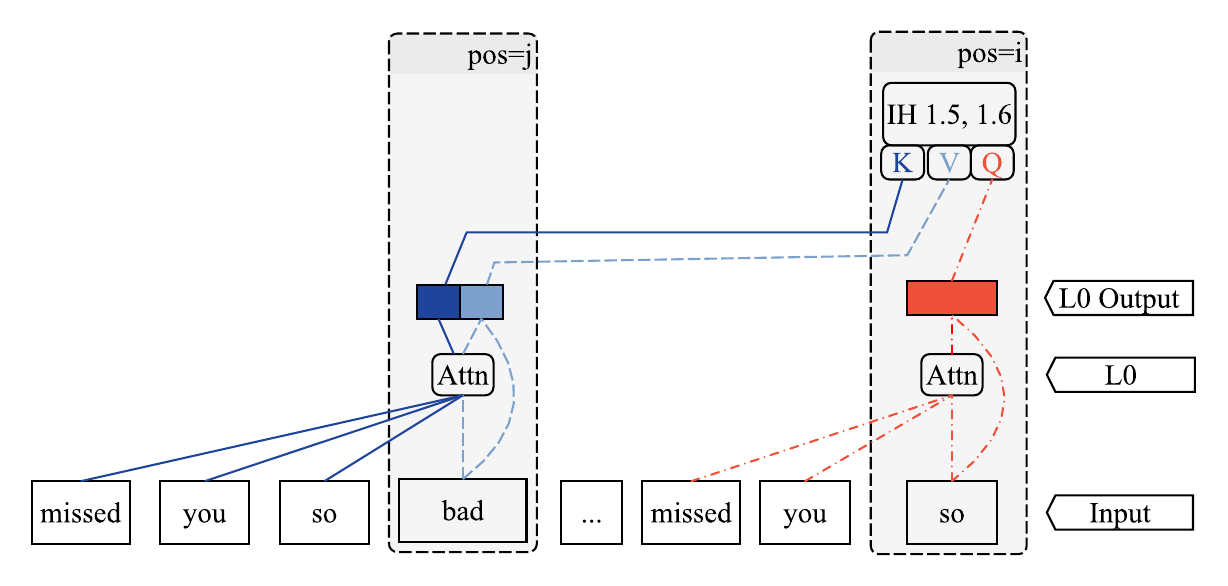}
\end{center}
\caption{the All-Long hypothesis with $K=3$}
\label{all_long}
\end{figure}

The results indicate that increasing the induction context is beneficial, but there is still room for improvement.

\begin{table}[ht]
\caption{Results of long hypotheses}
\label{results_long}
\begin{center}
\begin{tabular}{l l|l l}
\multicolumn{2}{l|}{Query and Key} &
\multicolumn{2}{l}{All} \\
\hline
Positional-Q + Positional-K & 52.5\% & All-Positional & 48.6\% \\
\hline
Long-QK & 59.5\% & All-Long & 55.2\%
\end{tabular}
\end{center}
\end{table}

\subsection{Third refinement: 1.5 and repeating entities}

At this point our hypothesis contains all of the pathways that should be important for induction. One cause of the remaining unexplained effect could be that these heads are not exclusively implementing the induction behavior, as observed by \citet{Olsson2022IncontextLA}.

\begin{figure}[ht]
\begin{center}
\includegraphics[width=\linewidth]{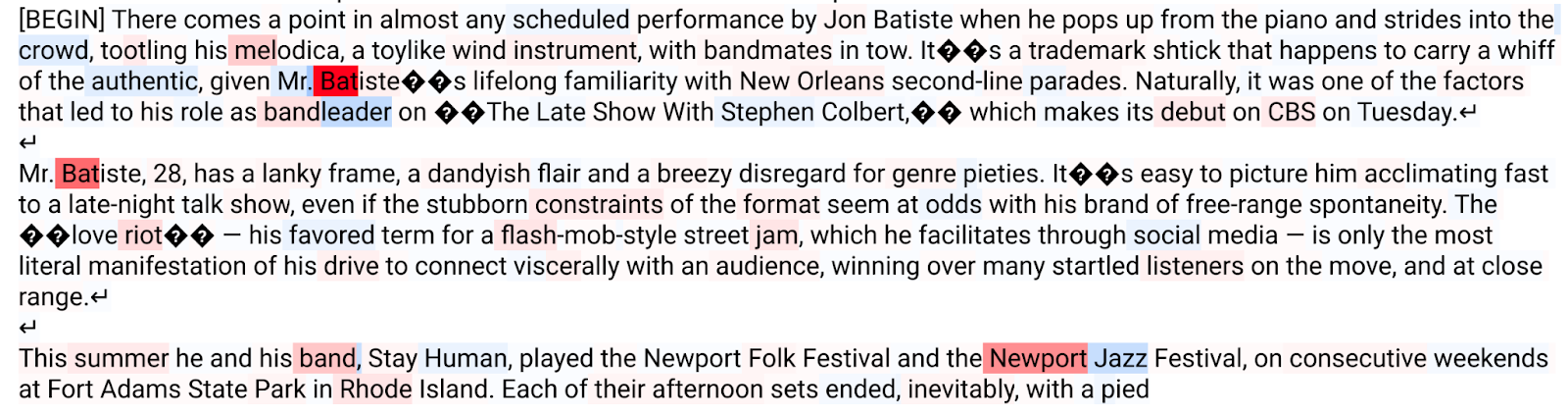}
\end{center}
\caption{Path patching attribution. Tokens with red/blue highlighting are where loss was higher/lower than the original model when all paths through 1.5 not included in All-Long are considered unimportant. For ``Newport Folk$\ldots$Newport" the original model predicts a second `` Folk" by induction, but the actual continuation is `` Jazz". The patched model is less capable of induction, so it gets lower loss on `` Jazz".}
\label{pp_attribution}
\end{figure}

In order to tell what these heads do that we might be missing, we'll use ``path patching attribution" \hyperref{sec:path_patching_attribution} to see where our hypothesis is unfaithful (Figure \ref{pp_attribution}).

The patched model tends to get higher loss on tokens like `` Bat" and `` Newport" when they have appeared previously. We propose that head 1.5 implements both induction and a distinct ``parroting" heuristic. Parroting says ``for each previous token, the fact that it appeared is evidence that it is likely to repeat". Mechanistically, parroting can be implemented by having 1.5's QK circuit attend to previous tokens in proportion to how likely they were to repeat on the training set. Attention due to induction and due to parroting are summed and softmaxed, and 1.5's OV copies over tokens in the appropriate ratios.

By defining narrower datasets, we can distinguish induction and parroting. We define an ``induction" subset where induction is helpful, and an ``repeats" subset, where copying a token from earlier in the context is helpful (but not necessarily because of induction). For more details see Appendix \ref{defining_subsets}. 

\begin{table}[ht]
\caption{All-Long results on subsets}
\label{results_subsets}
\begin{center}
\begin{tabular}{l |l |l |l}
& Full dataset & Induction subset & Repeats subset \\
\hline
AUE & 0.244 & 0.219 & 0.777 \\
\hline
Baseline (ATE) & 0.432 & 0.907 & 1.444

\end{tabular}
\end{center}
\end{table}

We do observe worse performance on the repeats subset. Inspecting the attention patterns (Appendix \ref{attn_15}) also gives evidence of attending to specific tokens in a context-independent manner. To successfully parrot, head 1.5 must be able to attend to $j$, so the paths $j$ $\rightarrow$ (L0 attn or skip) $\rightarrow$ head 1.5's K input are important. Adding these paths gives \emph{All-Final} with a proportion explained of 73.5\%, an increase of 18.3\% over All-Long and an increase of 45.3\% over All-Initial. 

In summary, while All-Initial is very sparse, it fails to capture most of the behavior. By adding a modest number of additional paths, we can obtain a much better approximation that gives insight into where the minimal story is insufficient.

\section{Path Patching vs Causal Tracing and Zero Ablation on GPT-2}

The causal tracing method of \citet{meng2022locating} can be considered a special case of path patching where the counterfactual input is sampled by adding Gaussian noise to the reference input. Another well-known technique is to zero ablate attention heads by simply skipping them in the computation (or equivalently, setting their output to zero). Either technique could be combined with our $\operatorname{Treeify}$ transformation, but we don't do this here.

In this section, we use a toy behavior to contrast the techniques and show that they can be used together to answer different questions. We give results on GPT-2 small (117M parameters) and GPT-2 XL (1.5B parameters).

Consider the set of prompts ``The organization estimates that [N]-" where N ranges from 0 to 100. When GPT-2 models are given these prompts, by inspection\footnote{Explore the behavior interactively at: \url{https://modelbehavior.ngrok.io/}} it appears that a combination of heuristics are used such as ``a number strictly larger but not hugely larger than N is likely" and ``round numbers are more likely, especially if N is round". 

Suppose we want to identify which attention heads use N to affect the output distribution. First, we run one experiment for each of the attention heads (144 in GPT-2 small) where all paths through that head are unimportant, and everything else (including MLPs) is important. We compute the mean KL divergence over 100 $(x_r, x_c)$ pairs. 

\begin{figure}[ht]
\begin{center}
\includegraphics[width=0.8\linewidth]{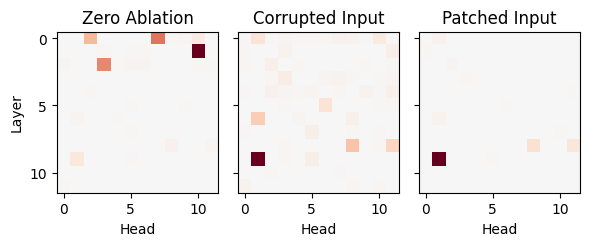}
\end{center}
\caption{GPT-2 small. For each head, darker color indicates a larger mean KL divergence from the original model when that head is ablated or patched. For casual tracing, we sample noise from $\mathcal{N}(0, 0.2)$.}
\begin{center}
\includegraphics[width=0.8\linewidth]{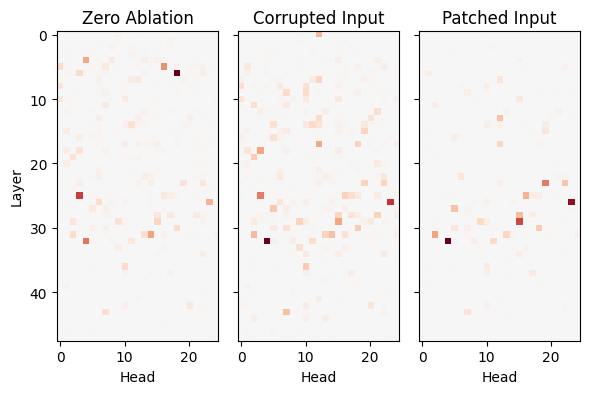}
\end{center}
\caption{As above, but for GPT-2 XL.}
\label{range_small}
\end{figure}

Path patching shows fewer heads than input corruption. One reason to expect this is since the counterfactual input is as similar as possible to the reference, the network's activations stay more on-distribution. In contrast, the corrupted version of the prompt number is in general off distribution and could represent a non-number or something the network has never encountered before. Similarly, if an attention head doesn't output a zero vector when on-distribution, later heads that read from that head will also be taken off distribution. However, without a ground truth it's impossible to conclusively say which of the plots is more representative of the true mechanism. 

We speculate that for GPT-2 small, heads 6.1 and 5.6 help recognize that the token before the hyphen is a number. If this is true, then corrupting the input to these heads would disrupt the heuristic and decrease the chance that the completion is a number, while patching a different number should not affect these heads. Table \ref{number_recognizers} shows evidence consistent with this speculation.

\begin{table}[ht]
\caption{Corrupting heads 5.6 and 6.1 decreases the probability of the completion being a number, while patching those heads has no 
 effect. The baseline probability of a numerical completion is 88.1\%.}
\label{number_recognizers}
\begin{center}
\begin{tabular}{l|l|l|l|l|l|l}

Head & \multicolumn{3}{l|}{Corrupted Input} & \multicolumn{3}{l}{Patched Input} \\
\hline
& \multirow{2}{*}{KL} & \multirow{2}{*}{KL given} & \multirow{2}{*}{Prob of } & \multirow{2}{*}{KL} & \multirow{2}{*}{KL given} & \multirow{2}{*}{Prob of} \\
&    & \# completion             & \# completion                 & & \# completion             & \# completion \\
\hline
5.6 & 0.025 & 0.006 & 82.5\% & 0.011 & 0.008 & 88.0\% \\
\hline
6.1 & 0.017 & 0.004 & 83.9\% & 0.003 & 0.002 & 88.3\% \\
\hline
All Heads & 2.328 & 0.883 & 26.9\% & 3.108 & 3.393 & 88.1\%
\end{tabular}
\end{center}
\end{table}

Much of the power of path patching is that it allows testing very specific claims: because we constructed a dataset where the counterfactual input is always a number, we are able to precisely exclude heads like 5.6 and 6.1 from consideration.

\section{Greedily building hypotheses}

Generation of hypotheses is currently labor intensive, and in the induction result we relied on visual inspection and domain knowledge. To reduce human labor, a naive automated baseline is to greedily add heads in descending order of mean KL. A greedy algorithm is not at all optimal, and more advanced techniques such as \citet{conmyacdc} should perform better, but this is relatively quick to run and hypotheses produced in this way can serve as a starting point for further refinement.

\begin{figure}[ht]
\begin{center}
\includegraphics[width=0.7\linewidth]{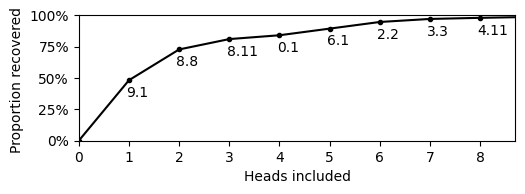}
\end{center}
\caption{GPT-2 small. The hypothesis ``paths through all heads are unimportant except 9.1 and 8.8" recovers 72.8\% of the loss. Including 8 of the 144 heads as important recovers 98.0\%.}
\includegraphics[width=\linewidth]{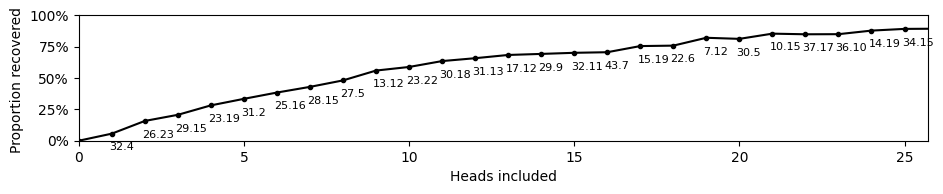}
\caption{As above, for GPT2-XL which has $48 \times 25 = 1200$ heads.}
\label{greedy_small}
\end{figure}

\section{Related Work}

\subsection{Causal Scrubbing}

Path patching is a simpler special case of causal scrubbing \citep{chan2022} where we make no claim about equivalence classes on nodes. Path patching experiments are more computationally efficient because only two samples $x_c$ and $x_r$ are needed, while in causal scrubbing you generally use more distinct samples.

\subsection{Causal Mediation Theory}

\subsection{Causal Mediation for Neural Networks}

\citet{wang2022interpretability} combine mean ablation with a simple form of path patching and identify a circuit of 26 attention heads that explain GPT-2 small's ability to identify indirect objects on a synthetic dataset. Their faithfulness metric is a difference of expectations, which is susceptible to the same possibility of cancellation described in \citet{scheurer2023}.

\citet{vig2020investigating} also apply causal mediation analysis \citep{pearl2022direct} to Transformer language models. While we measure the effect of altering unimportant nodes, they measure the effect of altering important nodes. For example, in the prompt ``The nurse said that", replacing ``nurse" with ``man" changes the completion from ``she" to ``he". They intervene on individual neurons or attention heads and do not consider paths.

\citet{geiger2020neural} apply interchange interventions to BERT on a lexical entailment task. Interchange interventions operate on nodes instead of paths, but they are more detailed in that they specify semantic meaning for the node such as ``this important node uses information about entailment". As with \citet{vig2020investigating}, while we intervene on the unimportant paths holding the important paths constant and expect no change, interchange interventions intervene on an important node and use the semantic meaning to predict how the model output should change.

\citet{geiger2021causal} apply interchange intervention to BERT on a more complex natural language inference task. They consider each example as a vertex in a graph, then adding an edge between examples a, b if the hypothesis holds both when intervening from a to b and from b to a. This identifies cliques in the graph for which the hypothesis holds fully. This avoids the complexity of measuring how approximate an abstraction is, at the cost of only applying to a very restricted set of inputs. 

\citet{finlayson2021causal} apply activation patching to study subject-verb agreement in GPT-2 and other transformers. They measure the relative probabilities of the correct and incorrect verb tenses; this means that heads only affecting other logits (or affecting correct and incorrect equally) don't have to be explained.

\citet{geiger2023causal} generalize causal abstraction, reframe existing methods into the causal abstraction framework, and give their own notion of approximate causal abstraction.

\section{Discussion}

In our induction head investigation we found that the prefix matching and copying behaviors are relatively separate phenomena, and in particular head 1.5 in our model performs copying both with and without prefix matching. It may be that the induction heads perform additional behaviors not characterized here. 

Compared to causal tracing and zero ablation, path patching is able to apply more precisely targeted interventions and thereby obtain a sparser abstraction.

\subsection{Limitations}

\paragraph*{Measuring sufficiency rather than completeness.} The AUE metric answers a very specific question: in expectation, how \emph{sufficient} is your chosen set of paths to mediate changes in output on the given distribution? For a simplified example, suppose your behavior is implemented by majority vote of a number of identical voters. Any set of important paths containing a majority of the voters will reach 0 AUE because their agreement is sufficient to mediate the output of the vote regardless of the minority's outputs. Of course, a set of paths containing all voters will also have 0 AUE, but since this hypothesis is less parsimonious we would by default not prefer it. 

A more realistic example would be that the voters represent correlated heuristics, whose weighted outputs feed into a saturating nonlinearity like sigmoid. Approximately the same thing happens: a set with most of the voting power will usually agree enough to saturate the sigmoid in either direction. Then this compact set achieves low AUE already and compares favorably to the set with all voters and marginally lower AUE. Therefore, it's important to interpret AUE-based metrics in terms of sufficiency and not completeness.

\paragraph*{No evidence about downstream tasks.} While this work is motivated by finding abstractions suitable for downstream tasks, in this work we don't establish the connection between AUE and specific applications. \citet{tay2022scale} found that for language models, pretraining perplexity is related to downstream performance, but that other factors like model shape also matter. Similarly, we expect AUE to be incomplete as a predictor of abstraction quality.

\paragraph*{Path patching makes no claims outside the tested distribution.}
 We consider paths that are approximately zero on the data distribution to be unimportant, as well as sets of paths that approximately cancel. The reason these contributions are small is often contingent on properties of the distribution, so low AUE doesn't imply low AUE on a wider distribution where those properties don't hold. This is comparable to training a model: low loss on some training distribution doesn't imply low loss off-distribution. Traditional techniques like an appropriate inductive bias, increasing dataset diversity, and use of regularization could increase the chances of generalization; other possible avenues are designing architectures to be easier to interpret in the first place or incorporating incentives for interpretability into the training process \citep{pmlr-v162-geiger22a, chen2019looks, elhage2022solu}

\paragraph*{Path patching cannot reject all false hypotheses.} Even if our dataset is the full distribution of interest, we can still fail to reject false hypotheses for reasons different than a too-narrow dataset:
\begin{itemize}
\item Some metrics, e.g. difference in expected loss, are susceptible to cancellation across examples: if the loss is higher on some examples and lower on others, these discrepancies can cancel. This is why we recommend using metrics that don't suffer from this, like average KL divergence.
\item AUE must be estimated from a finite number of samples in practice; if the computational budget is small relative to the diversity of the distribution, then naive sampling can fail to sample inputs that are rare but have a large unexplained effect.
\end{itemize}

\paragraph*{Path patching alone cannot definitively prove hypotheses.}

Given a correct hypothesis, proving it is correct with path patching would require testing all possible inputs, which is infeasible on large networks. We see path patching as providing evidence complementary with other lines of evidence like labor-intensive mechanistic analysis and other interpretability techniques.

\paragraph*{Narrow scope of demonstrations.}

Additional experiments are needed to characterize the effectiveness of the methodology on a diverse set of tasks and models. The results presented are narrow in scope compared to the full range of behaviors exhibited by language models.

\subsection{Future Work}

\paragraph*{Scaling To State of the Art Models.}
We have used path patching at the 1.5B parameter scale, but state of the art models are still orders of magnitude larger. Many interpretability techniques have only been applied to small networks \citep{rauker2023transparent}, and it remains to be shown that path patching scales to the largest models.

\paragraph*{Detecting Distribution Shift.}

One downstream application is the use of interpretability to detect distribution shift. Suppose that a given behavior is consistently explained by some hypothesis during training, but in deployment we observe that the same behavior is no longer explained by that hypothesis. Even if the actual outputs are indistinguishable, we could flag that the output is now produced for a different ``reason" and investigate the anomaly further.

\paragraph*{Producing Adversarial Examples.}

It should be possible to produce adversarial examples based on knowledge about model mechanisms. It would be useful to correlate the AUE metric with downstream performance on this task: how accurate do explanations need to be to produce adversarial examples in this way?

\paragraph*{Automating Hypothesis Search.}

Given a computational graph, it's straightforward to search over combinations of paths and automatically find candidates with favorable combinations of low AUE and sparseness. Ground truth labels provided by a tool like Tracr \citep{lindner2023tracr} could be used to benchmark search methods.

\subsection{Conclusion}

As neural networks continue to grow in capabilities, it is increasingly important to rigorously characterize their behavior. Path patching is an expressive formalism for localization claims that is both principled and sufficiently efficient to run on real models.


\subsubsection*{Acknowledgments}
We would like to thank Stephen Casper, Jenny Nitishinskaya, Nate Thomas, Fabien Roger, Kshitij Sachan, Buck Shlegeris, and Ben Toner for feedback on a draft of this paper.

\bibliography{iclr2021_conference}
\bibliographystyle{iclr2021_conference}

\appendix
\section{Why reuse the same counterfactual input?}
\label{reuse_xc}

\paragraph*{Reuse allows rewrites to preserve unimportance.}

If a node is unimportant and we rewrite it into any number of nodes that together add to the original node, then intuitively the new nodes considered together should also be unimportant. This can only be guaranteed by using the same $x_c$ for both new nodes. 

\paragraph*{Reuse does the right thing for additive ensembles.}

Residual networks often act like ensembles, where a behavior is implemented by combining contributions from multiple distributed components \citep{veit2016residual}. In particular, using dropout during training incentivizes this because then the behavior will degrade gradually instead of abruptly when a fraction of contributions are set to zero.

\begin{figure}[ht]
\begin{center}
\includegraphics[scale=0.5]{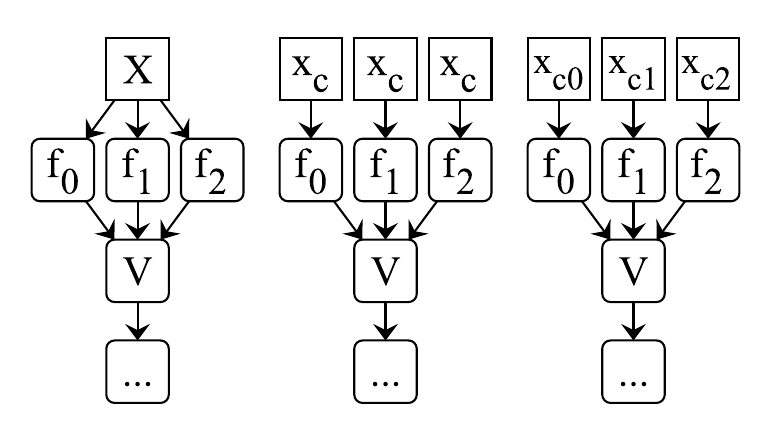}
\end{center}
\caption{Left: suppose $f_0$, $f_1$, and $f_2$ all compute the same function $f$ and then V takes the mean of its 3 inputs. Center: reusing the same $x_c$ means that V will output $f(x_c)$, which matches our intuition and is on distribution whenever $x_c$ is on distribution. Right: using distinct $x_ci$ means V outputs $(f(x_c1) + f(x_c2) + f(x_c3))/3$, which is off distribution in general and can cause unusual behavior later in the network.}
\label{ensembling}
\end{figure}

\paragraph*{Reuse allows paths that cancel across the domain to be unimportant.}

Suppose that in the figure below, $f_0(x) \approx -x$ on the domain $D_r$. When we sample $x_r \sim D_r, x_c \sim D_r$, testing each path individually rejects the claim of unimportance - clearly they each individually mediate the output.

\begin{figure}[ht]
\begin{center}
\includegraphics[scale=0.5]{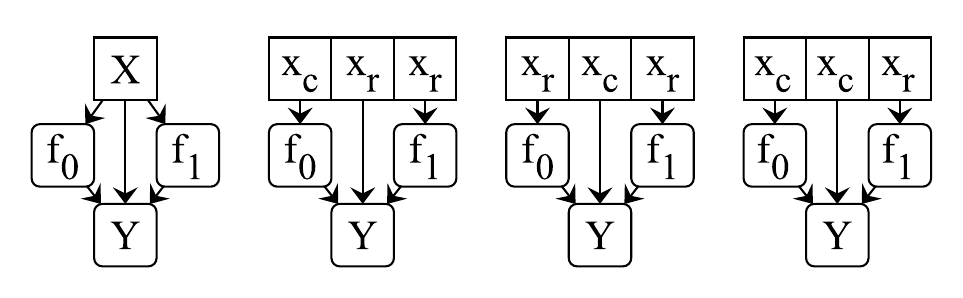}
\end{center}
\caption{For the graph on the left, tests that check if (a) $x \rightarrow f_0 \rightarrow Y$, (b) $x \rightarrow Y$, and (c) both together are unimportant.}
\label{cancellation}
\end{figure}

However, if we consider both paths together, they approximately cancel out - intuitively they are not important anymore. Concretely, this could happen if $f_0$ learns to erase information in $x$ that is irrelevant or harmful to predictions on $D_r$. Note that we can only exclude both paths from our explanation because we are restricting our claims to $D_r$ - we wouldn't expect this simplification to hold on arbitrary inputs.

\paragraph*{Reuse is computationally efficient.}

By reusing $x_c$, we are able to reuse sub-expressions that use $x_c$. For example, suppose our hypothesis for the above network was that was that only $x \rightarrow A \rightarrow f_1$ is important. To test this we evaluate: 

\[ G_T(x_c, x_c, x_r, x_c) = f_1(f_0(x_c) + x_r) + f_0(x_c) + x_c \]

In this case our software would compute the repeated term $f_0(x_c)$ only once and cache it; if the two terms were $f_0(x_{c0})$ and $f_0(x_{c1})$ this would not be possible. For a residual network, the number of paths grows exponentially in network depth and it's beneficial to cache as much as possible when working with large models.

\section{Identifying previous token and induction heads}
\label{identifying_heads}

Consider the string ``Mr. Dursley, Mrs. Dursley, Dudley Dursley" and let the number of tokens in the tokenized string be N. Because `` Dursley" is tokenized as `` D", ``urs", ``ley", we would expect an induction head to attend to the first occurrence of ``urs" from the second and third occurrences of `` D", the first occurrence of ``ley" from the second and third occurrences of ``urs", etc. 

For each attention head, we plot a NxN heatmap, where the i-th row represents the attention pattern over all tokens when the i-th token is the query and the j-th column represents the attention probabilities given to the j-th token as we vary the query token. Induction heads would then exhibit short diagonal patterns where the rows and columns are two distinct occurrences of the token sequence `` D", ``urs", ``ley".

\begin{figure}[ht]
\begin{center}
\includegraphics[width=\linewidth7]{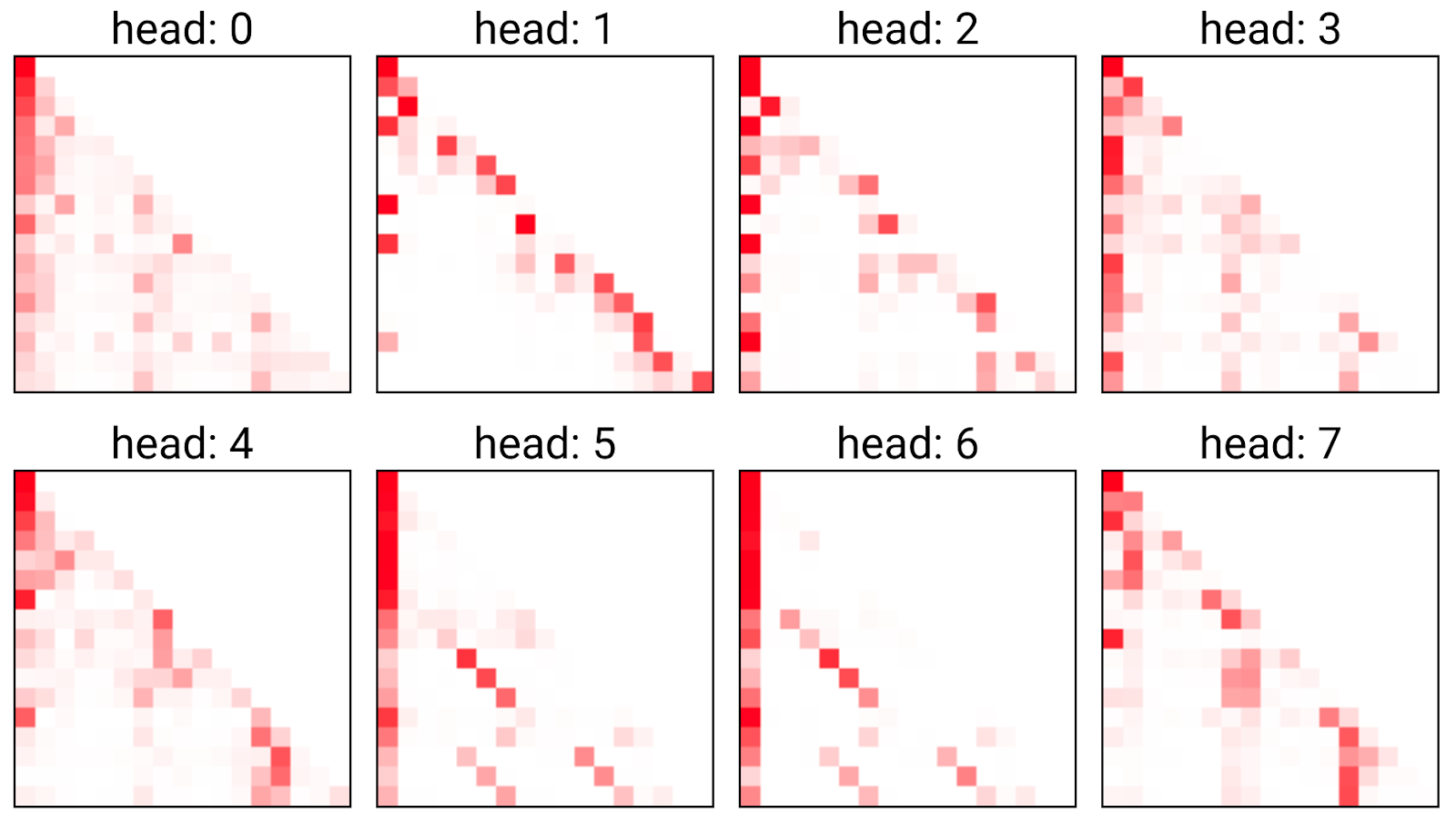}
\end{center}
\caption{Layer 1 heads. Heads 1.5 and 1.6 exhibit clear induction patterns.}
\label{l1_heads}
\end{figure}

To identify the previous-token head, we look at the corresponding attention pattern plots in the 0-th layer. The pattern of interest here is simply a reliable diagonal line running immediately below the main diagonal of the plot (for each query token, the previous-token head should attend primarily to the immediately preceding token). We find that head 0.0 fits the description:

\begin{figure}[ht]
\begin{center}
\includegraphics[width=\linewidth]{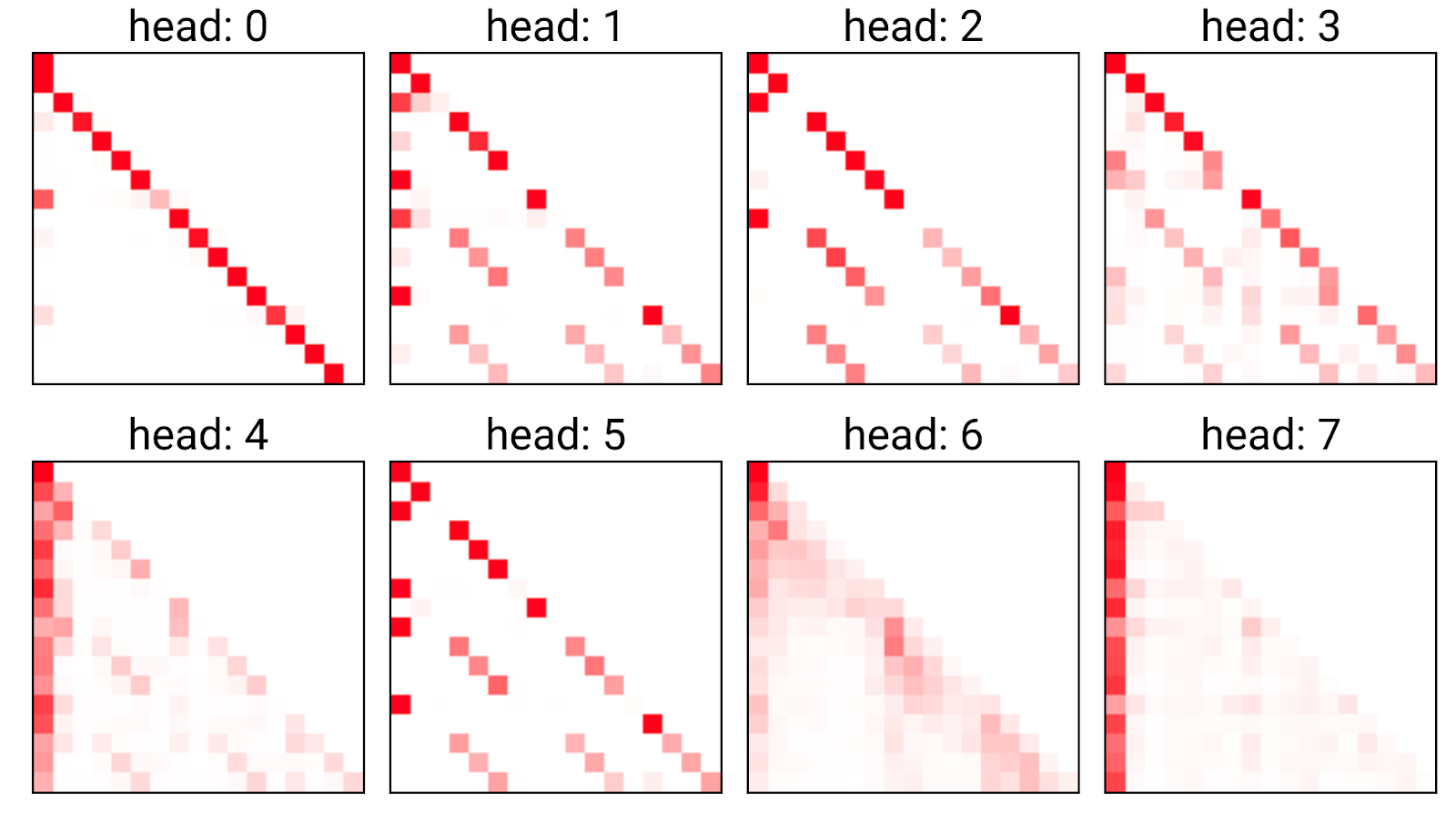}
\end{center}
\caption{Layer 0 heads. Head 0.0 is (primarily) a previous token head.}
\label{l0_heads}
\end{figure}

\section{Defining the induction and uncommon repeat subsets}
\label{defining_subsets}

The induction subset is the same as in \citet{chan2022}.

See \url{https://www.lesswrong.com/s/h95ayYYwMebGEYN5y/p/j6s9H9SHrEhEfuJnq#How_we_picked_the_subset_of_tokens} for a full description.

For the uncommon repeat subset, we include a token if it previously occurred in the context, and the token is not one of the 200 most common tokens in the validation set. We observed the results to be robust to varying the number of tokens filtered.

\section{Repeated Attention to Proper Nouns}
\label{attn_15}

\begin{center}
\label{ppattrib15}
\includegraphics[width=0.8\linewidth]{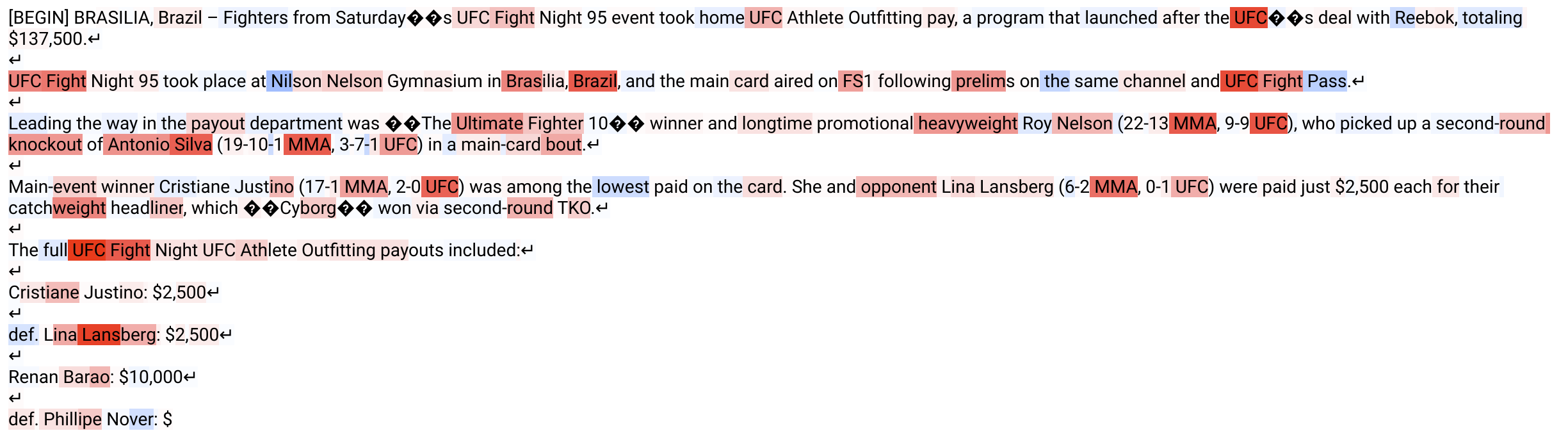}\\
Path patching attribution when patching head 1.5.
\end{center}

\begin{center}
\includegraphics[width=0.8\linewidth]{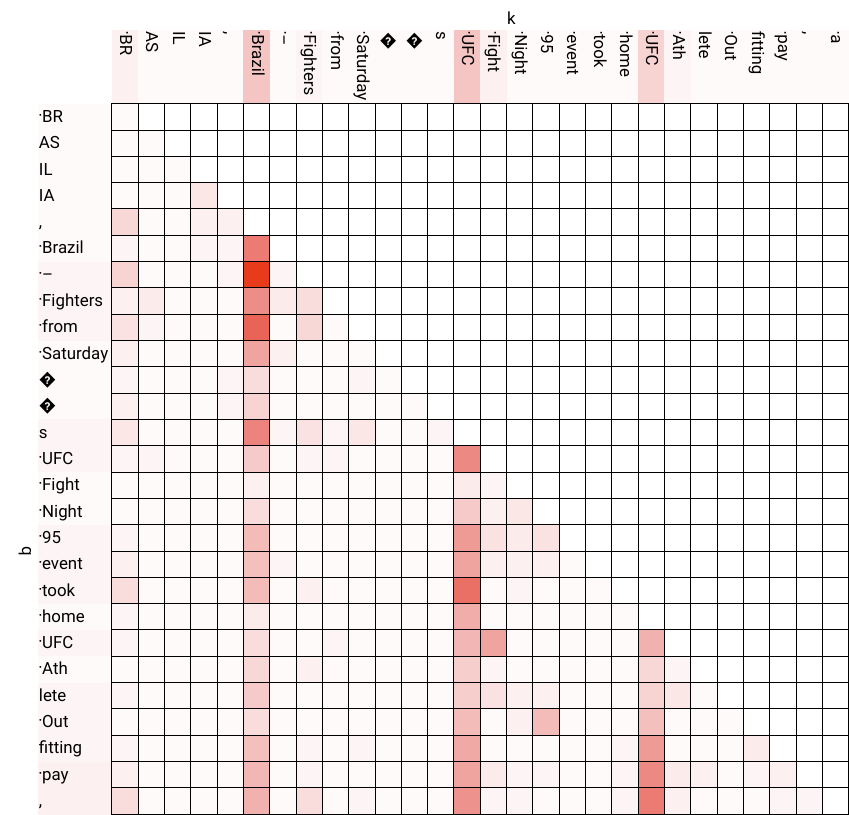} \\
\end{center}
The attention pattern of head 1.5 on an example that illustrates repeated attention to certain proper nouns such as `` Brazil" and `` UFC". Attention paid to the first token ``[BEGIN]" is not shown.

\section{Model Rewrites}
\label{model_rewrites}

Since the space of paths available depends on the exact structure of the computation graph, we often want to rewrite the graph to one that allows more fine-grained localization while preserving the behavior of $G$ on all inputs.

For example, we divide an attention layer into attention heads - now we can say that only some heads in a layer matter. 

Similarly, we can divide a vector of tokens into slices so we can say that only some tokens matter. 

The residual rewrite can be used to generalize subspaces and mean ablation.

Rewriting an activation as its projection onto a subspace and the remainder means you can check if each part is important individually.

Rewriting an activation into a constant value (specifically a mean) and the remainder. The constant is unimportant by definition (you can only resample it with the same constant), but then you can say the deviation is or isn't important.

\end{document}